\documentclass[lettersize,journal]{IEEEtran}

\usepackage{romannum}
\usepackage{xcolor}
\usepackage{multirow}
\usepackage{amsmath,amsfonts}
\usepackage{algorithmic}
\usepackage{algorithm}
\usepackage{array}
\usepackage[caption=false,font=normalsize,labelfont=sf,textfont=sf]{subfig}
\usepackage{textcomp}
\usepackage{stfloats}
\usepackage{url}
\usepackage{verbatim}
\usepackage{graphicx}
\usepackage{cite}
\usepackage{svg}
\begin{document}
%
\title{Damage Assessment after Natural Disasters with UAVs: Semantic Feature Extraction using Deep Learning}

\author{Nethmi S. Hewawiththi\textsuperscript{\ddag}, M. Mahesha Viduranga\textsuperscript{\ddag}, Vanodhya G. Warnasooriya, Tharindu Fernando,~\IEEEmembership{Member,~IEEE}, Himal A. Suraweera,~\IEEEmembership{Senior Member,~IEEE}, Sridha Sridharan,~\IEEEmembership{Life Senior Member,~IEEE}, and Clinton Fookes,~\IEEEmembership{Senior Member,~IEEE.}  

 \thanks{\ddag These authors contributed equally to this work.}
\thanks{N. S. Hewathithi and H. A. Suraweera are with Department of Electrical and Electronic Engineering, University of Peradeniya, Peradeniya, 20400, Sri Lanka (e-mail:  \{e17315, himal\}@eng.pdn.ac.lk).} 
\thanks{M. M. Viduranga is with the Multidisciplinary AI Research Centre, University of Peradeniya, Peradeniya, 20400, Sri Lanka (e-mail: e17360@eng.pdn.ac.lk).} 
\thanks{V. G. Warnasooriya is with Vega Innovations, Trace Expert City, Maradana, Colombo, Sri Lanka (e-mail: vanowarna@ieee.org).} 
\thanks{T. Fernando, S. Sridharan, and C. Fookes are with the Signal Processing, Artificial Intelligence, and Vision Technologies (SAIVT) research program at the Queensland University of Technology (QUT), Brisbane, Australia (e-mail: \{t.warnakulasuriya, s.sridharan, c.fookes\}@qut.edu.au).}
}






\maketitle
%



\begin{abstract}
Unmanned aerial vehicle-assisted disaster recovery missions have been promoted recently due to their reliability and flexibility. Machine learning algorithms running onboard significantly enhance the utility of UAVs by enabling real-time data processing and efficient decision-making, despite being in a resource-constrained environment. However, the limited bandwidth and intermittent connectivity make transmitting the outputs to ground stations challenging. This paper proposes a novel semantic extractor that can be adopted into any machine learning downstream task for identifying the critical data required for decision-making. The semantic extractor can be executed onboard which results in a reduction of data that needs to be transmitted to ground stations. We test the proposed architecture together with the semantic extractor on two publicly available datasets, FloodNet and RescueNet, for two downstream tasks: visual question answering and disaster damage level classification. Our experimental results demonstrate the proposed method maintains high accuracy across different downstream tasks while significantly reducing the volume of transmitted data, highlighting the effectiveness of our semantic extractor in capturing task-specific salient information.
\end{abstract}




\section{Introduction}
\IEEEPARstart {M}{ultiple} scientific studies show that hurricanes, heavy precipitation, and heatwaves are becoming more frequent in all parts of the world~\cite{WMO_2021}. Such extreme weather events lead to natural disasters and are associated with significant societal, economic, and environmental impacts, including the spread of infectious diseases, loss of human life, property damage, and increased pollution. According to the United Nations Office for Disaster Risk Reduction Global Assessment Report 2023~\cite{gar2023} which assumes a $2^\circ$C temperature surge due to climate change, riverine areas are projected to increase by $170\%$ relative to the period 1970-2020. With the rise of such disastrous events, effective disaster monitoring techniques are becoming increasingly important\cite{karaman2024solutions}. 

Unmanned aerial vehicles (UAVs) offer significant advantages over human disaster monitoring as they can monitor areas that are inaccessible to humans. UAVs are capable of monitoring disaster areas in real time, covering large areas in a short time, and providing real-time visual data to aid rescue teams in gaining situational awareness during emergencies~\cite{lou2016onboard,alsamhi2022uav, Ijaz_2023_UAV, Raja_2024_UGEN}. Fast and efficient communication of the images captured from UAVs to the rescue and evacuation teams is a vital component of enabling a real-time monitoring system. Communication delays could be a bottleneck in making timely decisions as there can be difficulties in finding a robust communication link during an emergency, resulting in numerous errors and considerable lags in transmitting the acquired information. Therefore, 
it is crucial to find a bandwidth-efficient mechanism for communication between UAVs and their associated ground stations to better leverage the advantages of UAVs during a crisis.  
\IEEEpubidadjcol
 
Semantic feature extraction \cite{Luo_WCM_2022} provides a mechanism through which the amount of data that needs to be transmitted from an aerial platform to a ground station can be reduced. Most of the existing state-of-the-art approaches in semantic extraction focus on either identifying context-based semantics that need to be transmitted or transmitting semantically segmented images where every pixel of the image is associated with a label. When transmitting context-based semantics, identifying the correct context of the captured data is of paramount importance, and incorrect interpretations of the context could result in miscommunication, misjudgments, and misguided decisions. Ensuring these systems can accurately interpret context across diverse scenarios can be challenging but it is vital in high-stakes environments such as disaster response, as the consequences of misinterpretations could be catastrophic. Most importantly, in existing state-of-the-art approaches\cite{lokumarambage2023wireless,semanticsegment_yu2018semantic,semanticmaskreiher2020sim2real} 
semantic information is transmitted to ground stations without considering the efficacy of the transmitted information to the intended application or the downstream task\cite{lokumarambage2023wireless, Lee_2024_UAVondevice}. This can be disadvantageous when considering different applications, as the type and amount of information needed differ depending on the application. 
  
To alleviate these issues, we propose a learnable semantics extractor where a trainable neural network, which can be trained for different downstream tasks, selects the information that needs to be transmitted. Specifically, this novel network component, which can be deployed onboard the UAV, can filter the application-specific information for transmission. Our proposed framework can dynamically adjust the information that it selects and transmits the most informative content for the downstream task at hand, which helps efficient decision-making with reduced bandwidth requirements for communication. 
  
In our approach, a semantic mask generation model first analyses the image and identifies the main object classes in the image. This step is followed by a binary mask predictor which is jointly trained with the downstream model. The binary mask is tasked with identifying the regions of the semantic mask which will be helpful in decision-making for the relevant downstream task. The data to be transmitted is formed by the product of a semantic mask with its binary mask, making the filtered data application-specific. In this way, the amount of information that needs to be transmitted is reduced while facilitating a robust decision-making process, which in turn results in a lesser response time in an emergency. Moreover, the developed framework is lightweight making it feasible to be executed in resource-constrained environments. 
The contribution of the paper can be summarized as follows:
  \begin{enumerate}
    \item We present a novel learnable semantics extractor that reduces transmitted data while preserving downstream task performance. 
    \item The proposed semantic extractor is task agnostic, making it a simple add-on to any existing architecture or downstream task. This adaptability is demonstrated with a diverse set of architectures and two applications, Image classification and Visual Question Answering (VQA) downstream tasks. 
    
    \item We demonstrate the utility of our novel semantic extractor on two real-world disaster monitoring datasets, namely, FloodNet \cite{Floodnet} and RescueNet \cite{Rescuenet}, where we achieve a significant reduction (i.e.$>$85\%) of the transmitted data while maintaining the performance in downstream tasks. 
  \end{enumerate}

This paper is organized as follows. Section \Romannum{2} summarizes the state-of-the-art approaches in using UAVs for disaster recovery missions and semantic extraction for efficient communication. Section \Romannum{3} presents the components of the proposed architecture in detail. Section \Romannum{4} describes the experiments conducted relevant to the proposed work and the results obtained through the experiments. Concluding comments are provided in Section \Romannum{5}.
%
%
%
%







\section{Related Works}
In this section, we discuss notable work to integrate UAVs into natural disaster recovery missions. Furthermore, this section provides an overview of the different approaches used for the extraction of semantics facilitating efficient communication.   
\subsection{Applications of UAVs in Natural Disaster Recovery Missions}
As UAVs are becoming faster, smaller, and more affordable, they are readily being adopted into a wide range of applications \cite{ Alotaibi_2019_SAR, Ezequiel_2014_UAVapplications}. Notably, UAVs have emerged as an effective technique where covering a geographical location is needed especially when it is out of reach of humans \cite{ Kyrkou_2024_UAV}.


Among this wide variety of applications of UAVs, natural disaster recovery missions take a major place due to their time criticality. Here, UAVs are deployed either during the natural disaster or after the disaster for escape, evacuation, and rescue assessment purposes \cite{Ijaz_2023_UAV, Dong_2021_SAR, Merkle_2023_DronesForGood}. This task involves not only capturing footage of the emergency area by UAVs but also efficiently extracting the information from the images for decision-making.

Therefore, the existing literature has focused on improving the efficiency of onboard processing to reduce the decision-making time in an emergency. 
For instance, the authors of \cite{alisjahbana2024deepdamagenet} propose to address the limitations in manual interpretation during building damage by using automatic classification by coupling a convolutional neural network (CNN)-based building identification model with a building damage classification CNN. An adaptive learning strategy in matching remote sensing images with a pre-trained model to improve classification accuracy is proposed in \cite{Wang_2022_RSimages}. 
An architecture named ``ERNet'' which can run fast on embedded platforms is proposed in \cite{Kyrkou_2019_ERNet}.
The methodology described in \cite{Rashid2024TinyVQACM} uses a two-layer early exit algorithm to maintain a balance between further processing and increasing accuracy for Visual Question Answering (VQA) tasks where the questions are related to a flooding event. The authors of \cite{Kyrkou_2020_EmergencyNet} suggest a lightweight convolutional neural network architecture for efficient aerial image classification. 

Moreover, a methodology to reduce training time and computational requirements in the VQA framework for onboard deployment is introduced in \cite{10126467}. The objective of this work is achieved by freezing the layers of the pre-trained encoder models and using the learned features without updating the model weights during training. In \cite{sarkar_chowdhury_murphy_gangopadhyay_rahnemoonfar_2023}, manually annotated visual masks are used for each question type and the gap between the attention matrix and the visual mask is reduced to increase the computational efficiency of the VQA task.  

A major weakness in the above methods is that these approaches are demonstrated for specific downstream tasks and it is unclear if they are flexible to be adopted for different downstream tasks.
Moreover, it is unclear whether architectures in these works of literature are applicable in limited bandwidth communication settings. We acknowledge that transmission of all captured information to a ground station where more communication bandwidth is available provides better knowledge extraction and decision-making with complex machine learning models. However, unlimited bandwidth for communication between the UAV and the ground station is impractical, especially in disasters as traditional terrestrial cellular networks could get disrupted during disasters by the failure of the base station infrastructure. As a solution for these limitations, we propose a bandwidth-effective communication method that learns to select application-specific critical information for decision-making such that it can efficiently transmit this information through limited communication bandwidth.
\subsection{Approaches for Semantic Extraction of Information}
In the process of selecting critical information, we concentrated on learning to extract semantics considering different downstream tasks. The current literature focuses on extracting contextual information from the images without knowing the downstream task or it purely transmits the semantic segmentation mask of the image \cite{9151332, 9756908}. In contrast, the proposed model is jointly trained with the downstream task, allowing the framework to dynamically determine which information is relevant to the downstream task. 
The earliest approaches leverage information theory to extract semantic information.
Another approach makes use of knowledge graphs to locate semantic information 
and some have utilized the power of modern deep-learning techniques to facilitate semantic feature extraction \cite{deepSCweng2021semantic}. 
Furthermore, some approaches focus on the significance of information, rather than its meaning, to achieve efficient communication. Another aspect named context-based semantic communication is also inspired by how humans naturally communicate with one another providing a generic, optimisation-based semantic communication system \cite{contextbasedsemanticzhang2022context}. The most common approach in semantic communication is to transmit semantically segmented images where every pixel of the image is associated with a label \cite{lokumarambage2023wireless}.

Despite these numerous approaches, one common drawback across all of these existing state-of-the-art methods is that the identified semantics in those methods are either application-specific or information reduction occurs without considering the importance of the transmitted information to the downstream task \cite{hu_wu_wu_xiong_2019}. This may lead to either the inclusion of non-critical information in the transmitted data which is an unnecessary transmission overhead, or missing the vital information that is needed for a particular downstream task. For instance, in a flood level assessment setting, the authors of \cite{https://doi.org/10.48550/arxiv.2407.05007} propose a method that selects information for transmission by identifying water and non-water areas of the captured image which might not be sufficient for decision-making in downstream tasks such as assessing the extent of the damage to the building. In contrast, our proposed approach can be adapted to any application without losing critical information since it automatically learns a task-specific critical from the data by jointly training with any downstream model. As such, our model is capable of reducing the data that needs to be transmitted, thereby the communication lag during transmission, by only transmitting the application-specific data for a particular downstream model, without sacrificing the accuracy of the downstream task. 

\begin{figure*}[htb]
\centering
\includegraphics[width=0.98\textwidth]{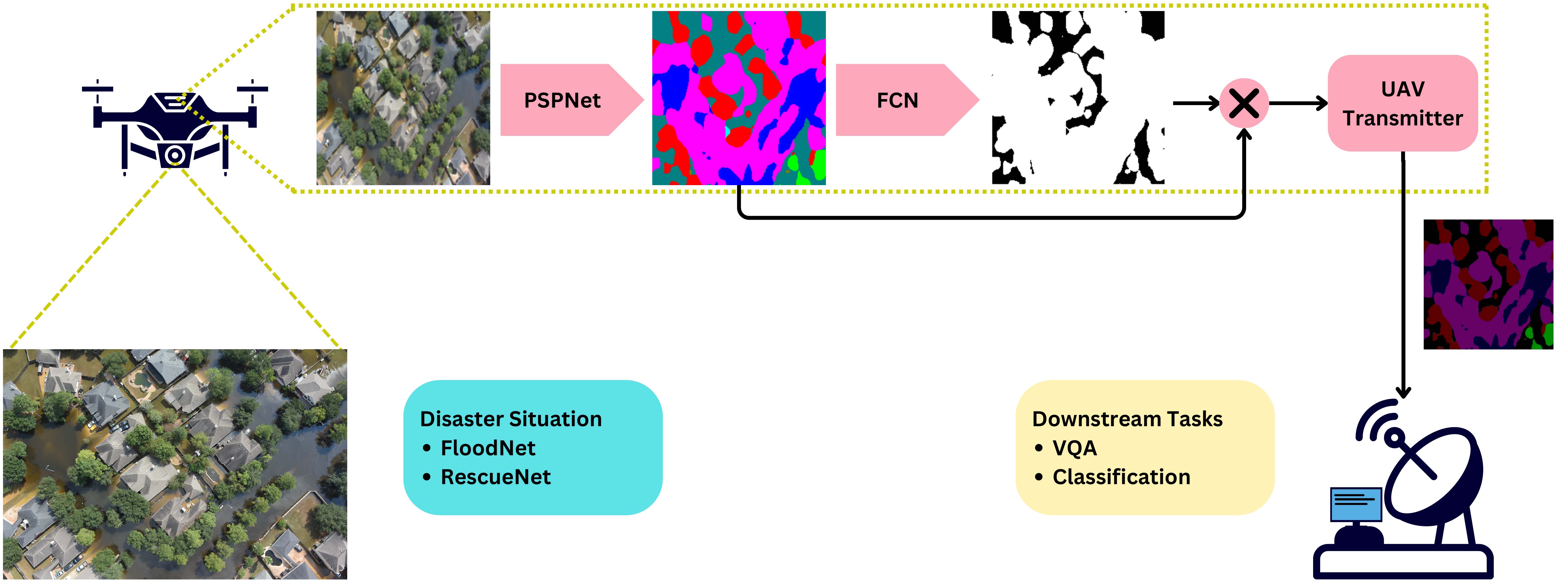}
\caption{An overview of the architecture of the proposed framework. First, the emergency area is captured using a UAV and it is converted into a semantic segmentation mask using PSPNet \cite{PSPNetzhao2017pyramid}. Next, a binary mask is created for the semantic segmentation mask using an FCN-based mask predictor to filter the critical data for the downstream task. Finally, the semantic mask is multiplied with the binary mask to filter unnecessary data making it ready to be transmitted to the ground station.}
\label{fig:overall_architecture}
\end{figure*}

\section{Methodology}
The overall architecture of the proposed model is shown in Fig. \ref{fig:overall_architecture}. First, the area of emergency is captured by a UAV, and the image is subjected to semantic segmentation based on PSPNet \cite{PSPNetzhao2017pyramid}. Next, salient image segments for the downstream task should be transmitted as selected by the proposed fully convolution neural network (FCN)-based binary mask predictor network. Finally, the predicted semantic mask from PSPNet and the predicted binary mask from the FCN model are multiplied to form the image that will be transmitted to the ground station. The following section discusses the semantic segmentation, data masking, and downstream task models in detail.  
\subsection{Semantic Segmentation Model}
The semantic segmentation model focuses on assigning a label for every image pixel which aids with the identification of key objects in the captured image. We adopt the PSPNet
\cite{PSPNetzhao2017pyramid} model for this task considering its performance and the feasibility of executing it onboard the UAV due to its efficiency. The architecture consists of two stages namely feature map generation and pyramid pooling module. In the feature map generation, the model generates the feature map using dilated convolutions. A pre-trained ResNet-50 architecture \cite{DBLP:journals/corr/HeZRS15} is used as the backbone for the extraction of feature maps. The extracted feature maps, \(F\), of size 1/8 times the input image can be expressed as,
\begin{equation}
F_{\frac{H}{8}\times\frac{W}{8}} = ResNet(I_{H\times W}),  
\end{equation}
where \(I_{H\times W}\) is the original image captured from UAV with height \(H\) and width \(W\). 

\begin{figure}[htb]
\centering
\includegraphics[width=0.49\textwidth]{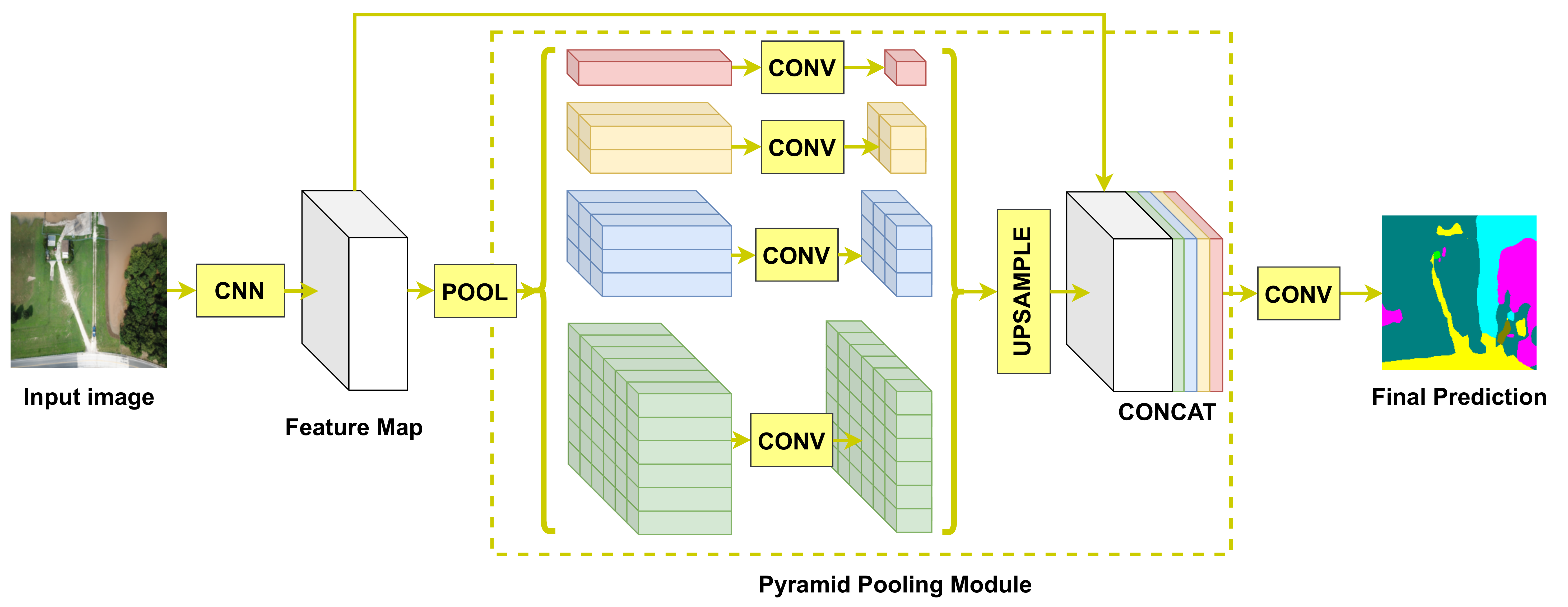}
\caption{Detailed structure of pyramid pooling module. First, the features are extracted at four different spatial scales through pyramid pooling. Here, the scales are 1 × 1, 2 × 2, 3 × 3, and 6 × 6. Then, 1 × 1 convolution is applied enhancing the nonlinear learning ability of the multiscale features. Next, a bilinear interpolation method is used to further interpolate convoluted feature maps. Finally, they are concatenated with the four upsampled feature maps.}
\label{fig:PSPNet}
\end{figure}

In the pyramid pooling module shown in Fig. \ref{fig:PSPNet}, feature maps are average pooled at different pool sizes for accurate segmentation of all the objects of different sizes. As in the original paper \cite{PSPNetzhao2017pyramid}, the four bin sizes that are used in the proposed model are $1\times1$, $2\times2$, $3\times3$, and $6\times6$. If the bin size is considered as $p\times p$, the feature map is divided into sub-regions of size $p\times p$ for each bin size. Feature map after average pooling $F_p$ can be obtained by, 
\begin{equation}
F_{p} = Pool_{p\times p}(F_{\frac{H}{8}\times\frac{W}{8}}),  
\end{equation}
where $Pool_{p\times p}$ is the average pooling performed on every $p\times p $ sub-region. 
The number of feature maps is reduced by applying $1\times 1$ convolution, which can be expressed as, 
\begin{equation}
F_{p,reduced} = Conv_{1\times 1}(F_{p}). 
\end{equation}
The reduced feature maps, $F_{p, reduced}$, are then up-sampled back to the size of the initial feature map using bi-linear interpolation, which is denoted as, 
\begin{equation}
F_{p,upsampled} = Upsample(F_{p,reduced}).  
\end{equation}
All the upsampled feature maps, $F_{p,upsampled}$, are concatenated with the original feature map, $F$, from the ResNet backbone as,
\begin{equation}
F_{p,concat} = Concat(F_{\frac{H}{8}\times\frac{W}{8}} ,F_{p,upsampled}).  
\end{equation}

The prediction of the semantic segmentation mask occurs after passing concatenated feature maps through a convolution layer represented as,
\begin{equation}
M = Conv(F_{p,concat}), 
\end{equation}
where $M$ is the predicted semantic mask. 


\subsection{Data Masking}
\begin{figure}[htb]
\centering
\includegraphics[width=0.5\textwidth]{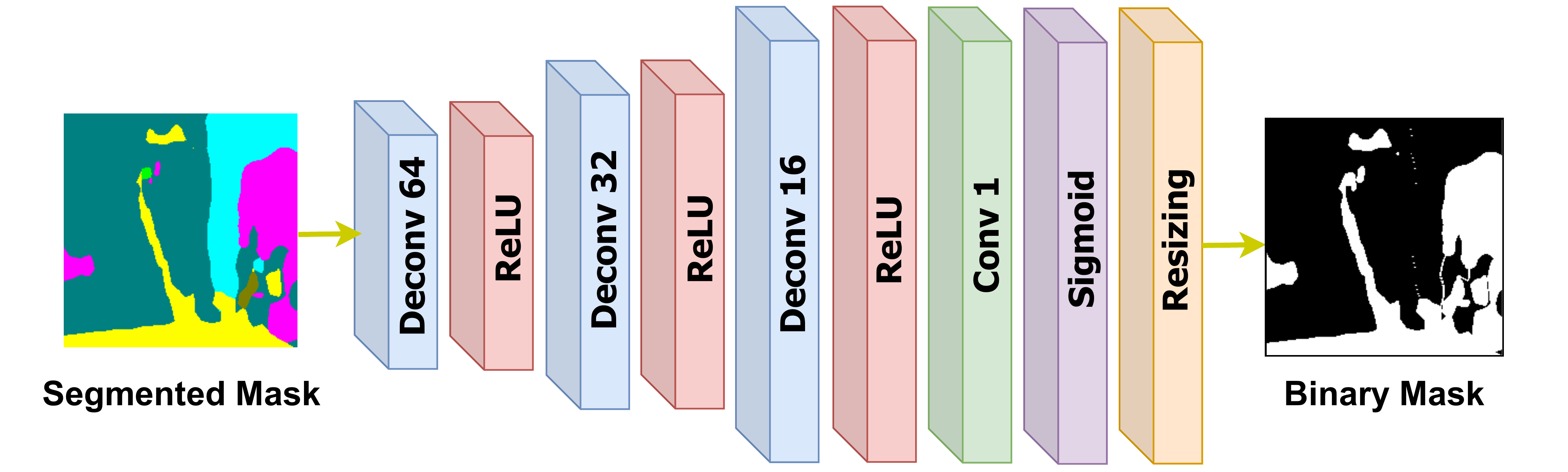}
\caption{The architecture of the data masking model. Initially, the segmented mask is sent through three transposed convolutional layers followed by a ReLU activation with an upsampling factor of two. Next, the output of the third transposed convolutional layer is sent via a convolution layer which is capable of creating a single channel output. Then, it is mapped into a binary mask between the values 0 and 1 through the Sigmoid function and finally resized to output a binary mask of the same size as the input image.}
\label{fig: mask FCN}
\end{figure}

Fig.\ref{fig: mask FCN} shows the architecture of the data masking model. The input for the binary mask predictor is the semantic segmentation mask, consisting of three channels.
Then, the image is sent through three transposed convolutional layers followed by a ReLU activation with an upsampling factor of two. Finally, a Gumbel-Softmax activation is applied to form a differentiable discrete choice of the transmitted information. The first transposed convolutional layer increases the spatial dimensions from 3 to 64 and the next two transposed convolutional layers reduce the spatial dimensions from 64 to 32 and from 32 to 16 respectively. 

Formally, the output of i\textsuperscript{th} transposed convolutional layer followed by a ReLU activation $f_i$ can be expressed as, 
\begin{equation}
f_i = ReLU(d_{4\times4}(f_{i-1})),  
\end{equation}
where \(d_{4 \times 4}(.)\) is \( 4 \times 4\) transposed convolution, \(ReLU(.)\) is ReLU activation and, 
\begin{equation}
f_0 = M.
\end{equation} 
The output of the third transposed convolutional layer is sent via a convolution layer reducing spatial dimensions from 16 to 1 creating a single channel output as expected. Furthermore, the Gumbel-Softmax function is applied to the output single-channel image for mapping the values between zero and one. 
The predicted binary mask $B$ for the image \(I\) can be denoted as,
\begin{equation}
B = \varphi(c_{1\times1}(f_3)),
\end{equation}
where \(c_{1\times1}(.)\) is \( 1 \times 1\) convolution and \(\varphi(.)\) is Gumbel-Softmax activation.

Afterward, the output is resized into the size of the predicted semantic mask 
such that the predicted semantic mask and the binary mask can be multiplied. The Gumbel-Softmax activation ensures that the process is differentiable. We need to enable binary mask prediction learning specific to a particular downstream model. As such, the training of the binary mask predictor occurs jointly with the downstream model.  This information is provided in Sec. \ref{sec:loss} of the manuscript. 
The image transmitted to the ground station (referred to as masked segmentation mask hereafter), $y$, can be obtained by, 
\begin{equation}
y = M \odot B,
\end{equation}
where \(\odot\) represents element-wise multiplication. 

\begin{figure}[htb]
\centering
\includegraphics[width=0.5\textwidth]{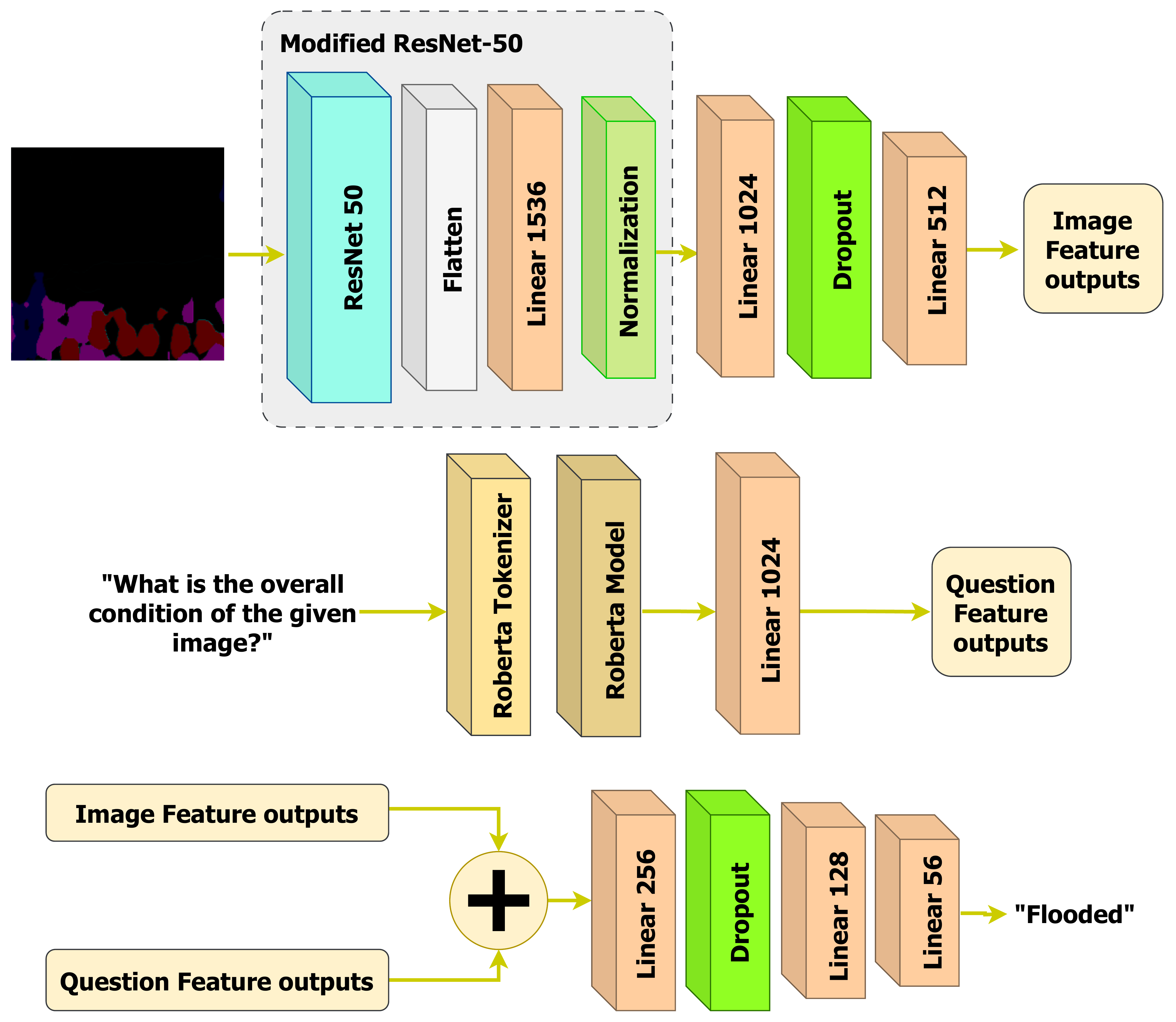}
\caption{The architecture of the visual question answering (VQA) model used in our work\cite{kane_khose_2022}. First, the masked segmentation mask and question features are created using visual and text encoders, respectively. Next, two features are fused using a combination mechanism and finally, an answer classifier is used to output the appropriate answer.}
\label{fig:VQA model}
\end{figure}

\subsection{Downstream Model}
The transmitted semantic information that has been extracted by the proposed framework can be used in different downstream models such as Visual Question Answering (VQA) related to a natural disaster or classification of the extent of the building damage. Furthermore, our presented pipeline can be trained for extracting semantic information to aid any other downstream task that the readers may be interested in. It should be noted that we selected VQA and classification as the two downstream tasks to demonstrate the diverse set of applications that our framework can facilitate. 

Since the downstream models are executed at ground stations with substantial computational resources, we develop large-scale models based on state-of-the-art techniques. These models are designed to achieve top-tier accuracy for the respective downstream tasks. The architecture of the downstream models for the VQA and classification tasks are described in the following subsections. 

\subsubsection{Visual Question Answering (VQA) Downstream Model}
One of the inputs to the VQA downstream model is the masked segmented mask, $y$, which is the output after the semantic mask and the binary mask are multiplied. The second input is the question that is asked for this input image. Inspired by \cite{kane_khose_2022}, our VQA model, presented in Fig.\ref{fig:VQA model}, combines these features with textual features and classifies them to predict the answer to the relevant question. To extract text features of questions we used a pre-trained transformer-based RoBERTa-Large model \cite{Liu2019RoBERTaAR}. Visual features are extracted using modified ResNet-50. Specifically, visual inputs are fed into two linear layers that reduce the dimensions of features and the dropout layer in the middle enhances regularization. Question features extracted from the RoBERTa model are passed through one linear layer with an output shape that is the same as the output shape of the visual feature model. Then both visual and question feature outputs are concatenated element-wise creating a combined vector. This combined vector further passed through a series of linear layers with one dropout layer. 
\begin{figure}[htb]
\centering
\includegraphics[width=0.5\textwidth]{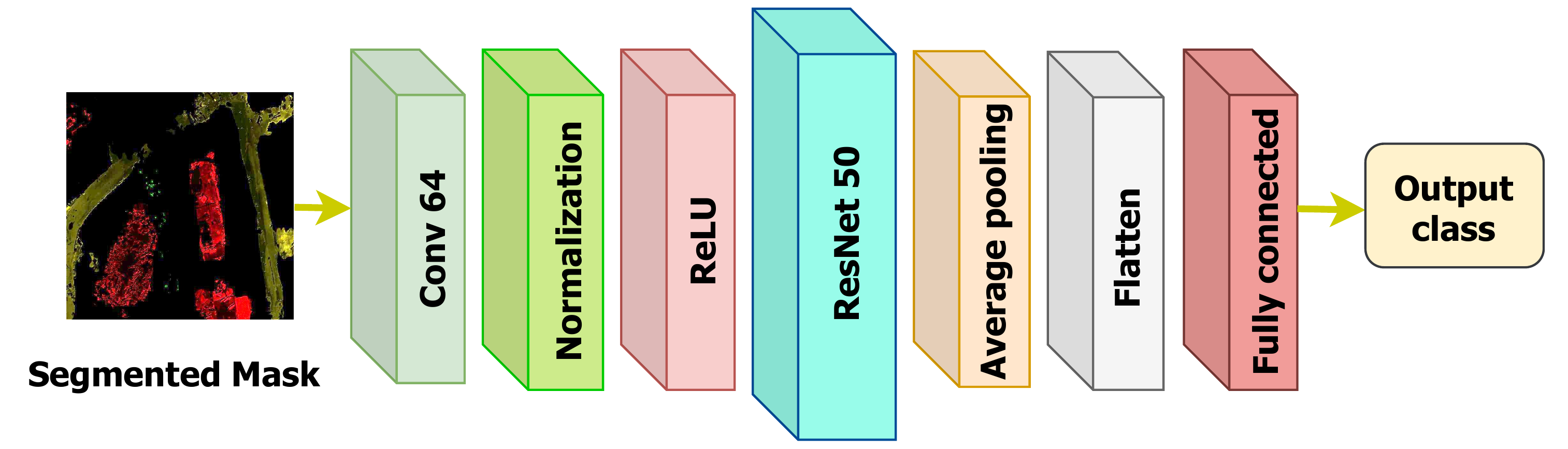}
\caption{The architecture of the proposed classifier model. First, the image transmitted to the ground station is sent through a convolutional layer followed by batch normalization and ReLU activation. Next, the data is consecutively passed through four main layers of the ResNet-50 backbone and an adaptive average pooling is applied to reduce the spatial dimensions of the feature maps into 1*1.  Finally, the output is flattened and a fully connected layer is used to prepare the final output of the model.}
\label{fig: classifier}
\end{figure}
The output dimension of the VQA model is 56 which is the same as the length of possible answers.
This output of the VQA model consists of the probabilities for each possible solution being the answer to a given question. The predicted answer, $\hat{a}$, for an image can be calculated by,
\begin{equation}
\hat{a} = \arg \max_{a \in A} f_\theta(a \mid x, Q),
\end{equation}
where $x$ is the VQA model, $Q$ is the question, $A$ is the possible set of answers and $f_\theta$ is the learnable model with $\theta$ parameters.

\subsubsection{Damage Extent Classification Model}

Fig.\ref{fig: classifier} shows the architecture of the proposed Damage Extent Classification Model which is our second downstream model. We leverage a ResNet-50 backbone for visual feature extraction, however, we do not utilize the pre-trained ImageNet weights as the input representations for our ResNet-50 backbone are significantly different from ImageNet. Prior to sending the inputs through the ResNet-50 backbone, we process the input with a single convolutional-batch normalisation-ReLu activation block that creates feature maps with 64 channels. The additional block before ResNet 50 enables us to have control over the parameters in the first convolution layer such as the number of input channels. Therefore, it facilitates changing the structure easily when working with augmented images.   Next, the data is passed through the first four layers of the ResNet-50 backbone to extract additional features. Subsequently, an adaptive average pooling is applied to the output to reduce the spatial dimensions of the feature maps into $1\times1$. Next, the data is prepared for the final fully connected layer by sending through a flattened layer and dropout is also applied to prevent overfitting. Finally, the data is sent through a fully connected layer to generate the damage extent classification.

\subsection{Loss Functions}\label{sec:loss}
Loss functions play a major role in the training process of each of the models mentioned above. Here, L1 loss is considered for the binary mask generating FCN model where the loss corresponds to the difference of the values of a particular binary mask and a matrix of zeros the same size. This loss encourages the selection of fewer segments from the semantic segmentation mask for transmission, thereby, reducing the size of the transmitted data. Calculation of the loss, $L_{sparcity}$ can be expressed as,
\begin{equation}
L_{sparcity} = \frac{1}{N} \sum_{i=1}^{N} |y_i - p|,
\end{equation}
where $L_{sparcity}$ is the L1 loss, $N$ is the number of training samples, $y_i$ is the predicted binary mask from the FCN for sample i, and $p$ is the matrix of zeros the same size. 

For training two downstream models: VQA and classification models, Categorical Cross Entropy Loss was considered for training purposes as each model engages in a multiclass classification task. Categorical cross-entropy loss is capable of measuring the difference between true distribution and predicted distribution calculated as,
\begin{equation}
L_{categorical} = -\frac{1}{N} \sum_{i=1}^{N}\sum_{c=1}^{\nu} \eta_c \log(\beta_c),
\end{equation}
where $L_{categorical}$ is the cross-entropy loss, $\nu$ is the number of classes, $\eta_c$ is the true label for class $c$, and $\beta_c$ is the predicted probability of the model classifying the sample as class $c$.

To generate the total loss of our framework the $L_{sparcity}$ loss of the binary mask predictor and relevant downstream task loss are combined with appropriate weights. The loss for training the binary mask predicting model $Loss$ can be calculated by
\begin{equation}
Loss = w_{s}L_{sparcity} + w_{c}L_{categorical},
\end{equation}
where $w_{s}$ and $w_{c}$ are loss weights which are experimentally evaluated.

\section{Experiments}
In this section, we present the experiments that we conducted to demonstrate the proposed model. The VQA downstream task is demonstrated using the FloodNet \cite{Floodnet} dataset and the damage extent classification task is demonstrated using the RescueNet \cite{Rescuenet} dataset. In addition to these experiments, which demonstrate that the proposed method can be adapted to numerous downstream tasks, we conduct additional evaluations to show our data masking model is light-weight making it feasible to be deployed onbard of a UAV. 
In the following subsections, we first introduce the datasets, evaluation protocol, and results.

\subsection{Datasets}
\begin{enumerate}[]
    \item FloodNet Dataset
    
    FloodNet dataset provides high-resolution \cite{Floodnet}  images taken from low altitudes using a UAV under flood conditions after Hurricane Harvey.
    It also offers pixel-level annotation for semantic segmentation and VQA tasks. For semantic segmentation, the images are annotated with 10 classes which include background, building-flooded, building-non-flooded, road-flooded, road-non-flooded, water, tree, vehicle, pool, and grass. For the VQA task, questions are divided into a five-way question group, namely ``Simple Counting'', ``Complex Counting'', ``yes/no'', ``Condition Recognition'', and ``Image Condition''. These questions have been designed to provide useful insights to end users such as the number of buildings flooded and whether a particular road is flooded. We used 1448 images of the FloodNet dataset, each with a size of 3000 by 4000 along with 4511 question-answer pairs. To make the VQA task challenging we assigned multiple question-answer pairs to the same image. The dataset was split into 60\%, 20\%, and 20\% percentages for train, validation, and test, respectively. 
    
    \item RescueNet Dataset\cite{Rescuenet}
    
    The RescueNet dataset is a high-resolution aerial imagery dataset captured using UAVs. It consists of both affected and non-affected areas following Hurricane Michael which affected Florida Panhandle in the United States with 3595, 449, and 450 images of size 3000 by 4000 for training, validation, and testing, respectively. It also offers pixel-level annotation for 10 classes including Water, building no damage, building minor damage, building major damage, building total destruction, vehicle, road-clear, road-blocked, tree, and pool. Moreover, each image in the dataset is also classified into one of three classes: Superficial damage, Medium damage, and Major damage. If the image does not exhibit any damaged structures, it is classified as ``Superficial damage''. In the case where a few structures are damaged by the natural disaster, the image is labeled as ``Medium damage''. Finally, if the image contains at least one completely destroyed building or if approximately 50\% of the area is covered with debris, it is categorized as ``Major damage''. 
  
\end{enumerate}

\subsection{Evaluation Protocol}
To better demonstrate the strengths of the proposed method we evaluated its error rate in the two downstream tasks, the amount of data that has to be transmitted from the UAV to the ground station, and the expected latency during transmission. Specifically, when quantitatively assessing the performance of the proposed model, the error rate is compared for the two downstream tasks using the original image, semantically segmented mask, and masked segmentation mask. Furthermore, to demonstrate the tradeoff between the amount of data transmitted and the accuracy of the downstream model, the transmission latency of the original image, semantically segmented mask, and the masked segmentation mask are also compared in a fixed bandwidth environment which mimics the transmission of the data from the UAV to the Base Station (BS). 

\textit{Transmission Latency Calculation:}
When calculating the latency in the UAV-to-ground BS link, it is assumed that a line-of-sight (LoS) path exists between the UAV and ground BS and therefore, the channel attenuation mainly depends on the UAV-to-ground BS distance~\cite{qi2024minimizing}. Consequently, using the free-space path loss model \cite{ khan2024efficient}, an expression for the achievable rate of the UAV-to-ground BS link can be established as,
\begin{equation}
R_b = \log_2 \left( 1 + \frac{P_u \left(\frac{\alpha_0}{B N_0}\right)}{(H_b - H_u)^2 + \|\mathbf{W}_b - \mathbf{q}_u\|^2} \right),
\end{equation}
where $P_u$ is the UAV transmit power, $\alpha_0$ is the channel gain at a reference distance of 1 m, $B$ is the bandwidth, $N_0$ is the noise power density, $H_u$ is the UAV elevation, $\mathbf{W}_b$ is the 2D coordinates of the ground BS and $\mathbf{q}_u$ is the 2D coordinates of UAV. 
As in \cite{qi2024minimizing}, the transmission latency, $t$, can be obtained from 
\begin{equation}
t = \frac{S}{R_b},
\end{equation}
where $S$ is the data size expected to be transmitted.  

\textit{Jaccard Index Calculation:}
The semantic fidelity of the features extracted by the model when the segmented mask and masked segmentation mask are fed is also evaluated by calculating the Jaccard index between the extracted features. This metric is capable of demonstrating how much semantically important data is extracted when the data filtering occurs. Jaccard index, $Jac(A,B)$, between $A$ and $B$  can be calculated as,
\begin{equation}
\text{Jac}(A, B) = \frac{|A \cap B|}{|A \cup B|},
\end{equation}
where $A$ and $B$ are two given sets.





\subsection{Results}
In this subsection, we provide both quantitative and qualitative results using the above specified protocols.  

Table \ref{tab:errorrate_vqa} summarizes the average error rate that occurred in different question types of the FloodNet dataset when the VQA model receives the original image, ground truth segmentation mask, predicted semantic mask, and masked segmentation mask (i.e. output of the proposed framework) as the inputs. Similarly, the average error rates of the damage extent classification task of the RescueNet dataset are provided in Table \ref{tab:errorrate_classification}. 

\begin{table*}
    \centering
    \caption{Average error rates for the VQA task of the FloodNet dataset. We provide evaluations when the original image, ground truth segmentation mask, predicted semantic mask, and the masked segmentation mask (i.e. output of the proposed framework) are used as the inputs to the VQA model.}
    \begin{tabular}{|c|c|c|c|c|}
        \hline
           & Original Image(\%) & Ground Truth Mask(\%) & Predicted Semantic Mask(\%) & Masked segmentation Mask(\%) \\
           \hline
        Overall & 31.11 & 30.67 & 33.19 & 31.00 \\
        \hline
        Simple Count & 71.77 & 74.19 & 80.64 & 76.61 \\
        \hline
        Complex Count & 74.82 & 75.54  & 75.54  & 76.26  \\
        \hline
        Road Condition & 3.33  & 3.33 & 7.22 & 9.44 \\
        \hline
        Yes/No & 38.89 & 35.00 & 35.00 & 24.44 \\
        \hline
        Image Condition & 5.17 & 4.83 & 7.59 & 7.24 \\
        \hline
    \end{tabular}
    
    \label{tab:errorrate_vqa}
\end{table*}

\begin{table*}
    \centering
    \caption{Average error rates for the building damage classification task of the RescueNet dataset. We provide evaluations when the original image, ground truth segmentation mask, predicted semantic mask, and the masked segmentation mask (i.e. output of the proposed framework) are used as the inputs to the Damage Extent Classification model.}
    \begin{tabular}{|c|c|c|c|c|}
        \hline
           & Original Image(\%) & Ground Truth Mask(\%) & Predicted Semantic Mask(\%) & Masked Semantic Mask(\%) \\
         \hline
        Classification Error Rate & 30.00 & 32.89 & 36.00 & 41.33 \\
        \hline
    \end{tabular}
    
    \label{tab:errorrate_classification}
\end{table*}

When analysing the results presented in Tables \ref{tab:errorrate_vqa} and \ref{tab:errorrate_classification} it is clear that the proposed data masking strategy does not significantly increase the error rates in the two downstream tasks. Furthermore, in the VQA task for question types such as ``Yes/No'' we observe a reduction in error rates when inputting the proposed masked segmentation mask compared to inputting the original image. This demonstrates that the filtering of unnecessary information for simple question types such as ``Yes/No'' questions would help with decision-making. Similarly, we would like to illustrate that for the building damage classification task the increase in the error from inputting the predicted semantic mask to the proposed masked segmentation mask is not significant. 

Moreover, the average data size of the original image, predicted segmentation mask, and masked segmentation mask, and the possible latency during transmission of those data are evaluated for the test data in the FloodNet and RescueNet datasets and the results are provided in Tables \ref{tab:Floodnet_processing} and  \ref{tab:Rescuenet_processing}, respectively.

\begin{figure*}[!htbp]
\centering
\includegraphics[width=1\textwidth]{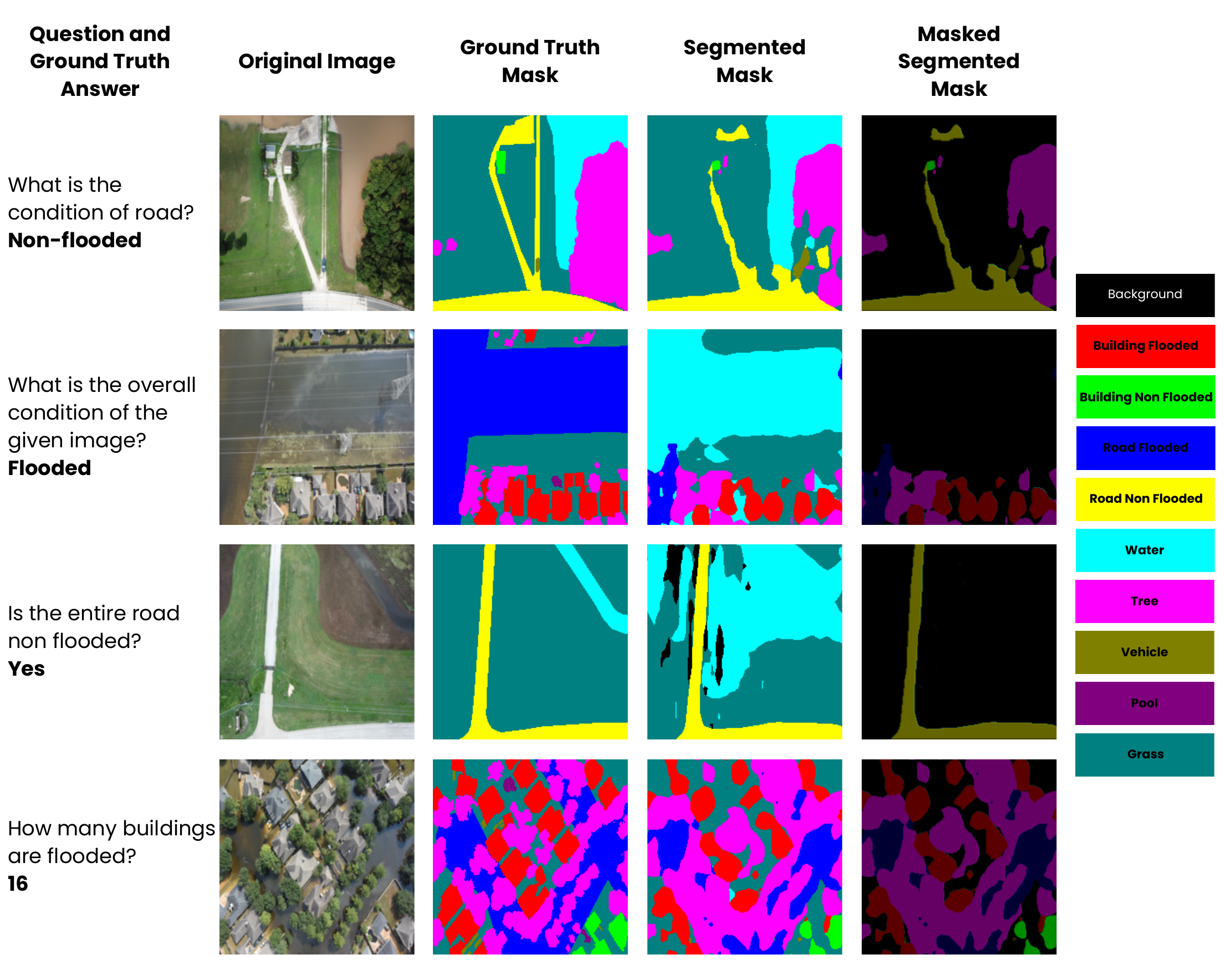}
\caption{Samples visualisations of the original image, ground segmentation truth mask, predicted segmentation mask, and masked segmentation mask for the VQA task using the FloodNet dataset.}
\label{fig: vqa_results}
\end{figure*}

\begin{figure*}[htb]
\centering
\includegraphics[width=1\textwidth]{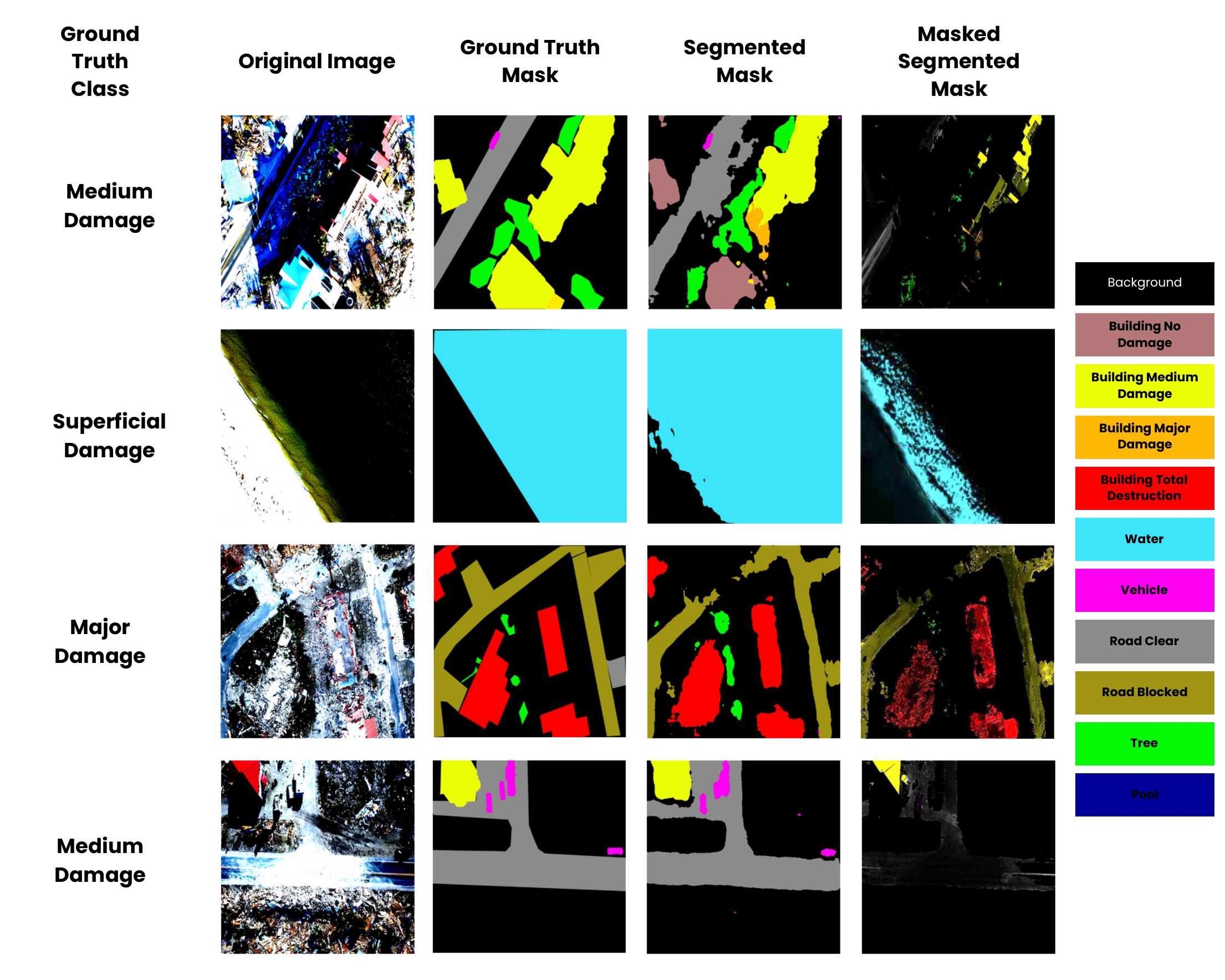}
\caption{Samples visualisations of the original image, ground segmentation truth mask, predicted segmentation mask, and masked segmentation mask for the damage extent classification task using the RescueNet dataset.}
\label{fig: classification_results}
\end{figure*}

\begin{figure*}[!htbp]
\begin{center}
\includegraphics[width=0.8\textwidth]{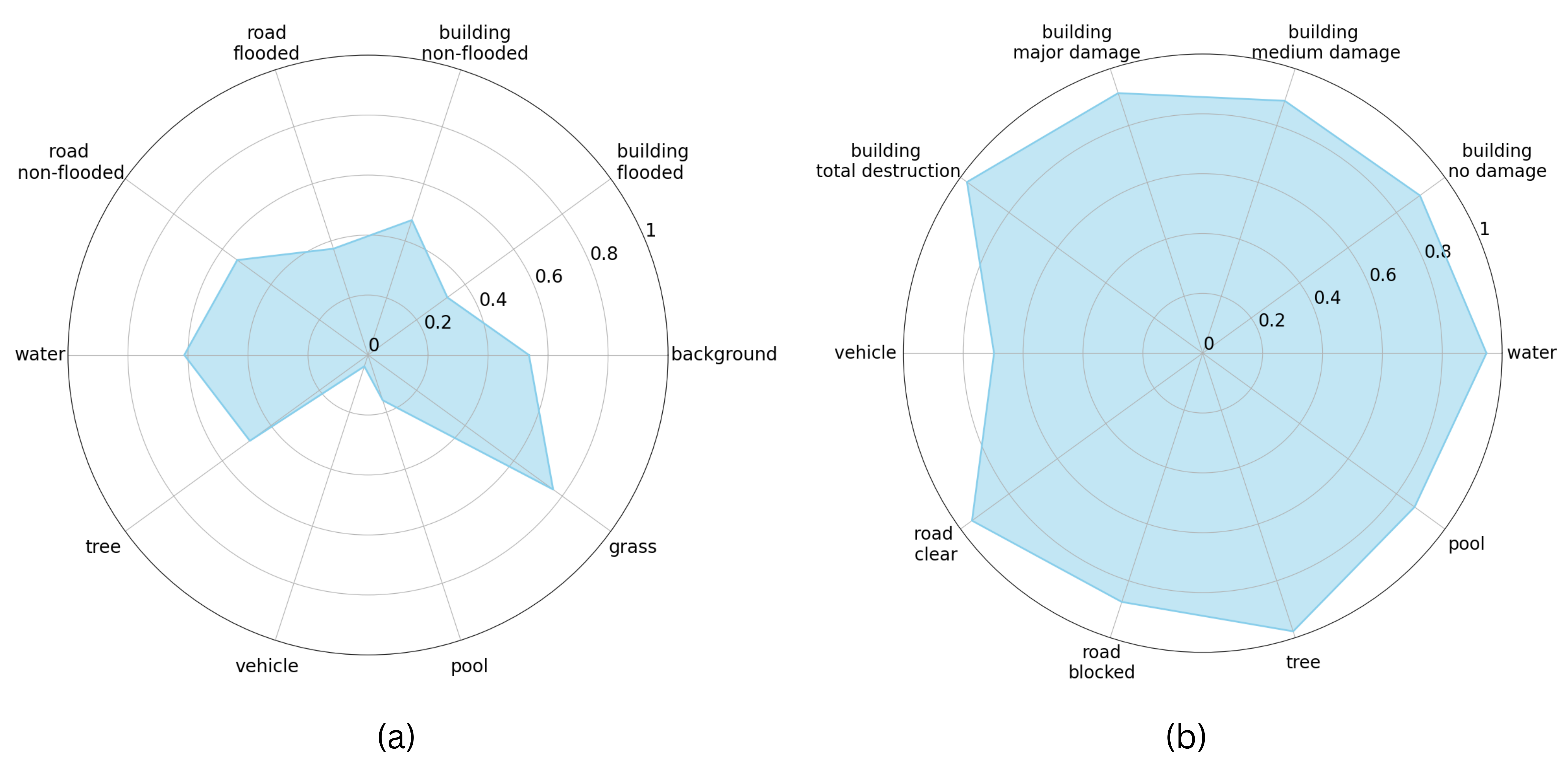}
\caption{Class wise mIoU on 0 to 1 scale between predicted masks using PSPNet and ground truth masks for datasets (a) Floodnet. (b) Rescuenet.}
\label{fig: class_iou}
\end{center}
\end{figure*}

Table \ref{tab:Floodnet_processing} and Table \ref{tab:Rescuenet_processing} show the evaluated transmission latencies for the VQA task and classification task using the FloodNet dataset and RescueNet dataset under four different UAV elevations. We set $P_u$ = 0.1 W, $\alpha_0$ = -60 dB, $B$ = 1 MHz, $N_0$ = -174 dBm/Hz, $H_b$ = 20 m, $W_b$ = $[-2500,0]^T$, and $q_u$ = $[0,1000]^T$ \cite{khan2024efficient}. According to the values given in Tables \ref{tab:Floodnet_processing} and \ref{tab:Rescuenet_processing}, the proposed binary masking strategy had caused approximately 86\% and 92\% reduction of the data size which needs to be transmitted as compared to the original images in FloodNet and RescueNet datasets, respectively. Hence, the latency in the transmission channel is expected to reduce significantly when transmitting a series of images at a very limited cost of accuracy. This is a major advantage in a disaster recovery mission as the early responses facilitated by faster transmission can save lives. Therefore, we argue that the small increment in error rate due to the absence of fine-grained details in the transmission data is negligible compared to the significant reduction in transmission latency that might occur when transmitting hundreds of images. 

Fig.\ref{fig: vqa_results} shows qualitative results of the original image, ground truth segmentation mask, predicted segmented mask, and masked segmentation mask for some sample questions in the FloodNet dataset. This visually illustrates how the elementwise multiplication of the predicted binary mask with the semantic mask filters the unnecessary data for different question types. Similarly, Fig \ref{fig: classification_results} shows the original image, ground truth mask, predicted segmented mask, and masked segmentation mask for samples in the RescueNet dataset. 
To quantitatively evaluate the performance of the PSPNet model on FloodNet and RescueNet datasets for the semantic segmentation task, the mean Intersection of Union (mIoU) is calculated separately for both datasets. Fig. \ref{fig: class_iou} graphically represents the classwise results. 


    

    

\begin{table*}[]
\centering
\caption{Performance Comparison for FloodNet Dataset}
\begin{tabular}{|>{\centering\arraybackslash}p{0.18\textwidth}|>{\centering\arraybackslash}p{0.18\textwidth}|>{\centering\arraybackslash}p{0.12\textwidth}|>{\centering\arraybackslash}p{0.12\textwidth}|>{\centering\arraybackslash}p{0.12\textwidth}|>{\centering\arraybackslash}p{0.12\textwidth}|}
\hline
\multirow{2}{*}{Method} & \multirow{2}{*}{\begin{tabular}[c]{@{}c@{}}Average Transmitted \\ Data Size (kB)\end{tabular}} & \multicolumn{4}{c|}{Transmission Latency (ms)} \\ \cline{3-6} 
                        &                                                                                               & Hu = 100 m          & Hu = 150 m         & Hu = 200 m         & Hu = 250 m        \\ \hline
Full Image              & 14.441                                                                                        &    6.692                &     6.698               &       6.704             &    6.717               \\ \hline
Ground Truth Mask              & 7.004                                                                                        &    3.246                &     3.249               &       3.251             &    3.258               \\ \hline
Segmented Mask          & 2.119                                                                                         &      0.982              &     0.983               &      0.984              &      0.986             \\ \hline
Masked Segmentation Mask& 1.945                                                                                         &      0.901              &     0.902               &     0.903               &     0.905              \\ \hline
\end{tabular}
\label{tab:Floodnet_processing}
\end{table*}

\begin{table*}[]
\centering
\caption{Performance Comparison for RescueNet Dataset}
\begin{tabular}{|>{\centering\arraybackslash}p{0.18\textwidth}|>{\centering\arraybackslash}p{0.18\textwidth}|>{\centering\arraybackslash}p{0.12\textwidth}|>{\centering\arraybackslash}p{0.12\textwidth}|>{\centering\arraybackslash}p{0.12\textwidth}|>{\centering\arraybackslash}p{0.12\textwidth}|}
\hline
\multirow{2}{*}{Method} & \multirow{2}{*}{\begin{tabular}[c]{@{}c@{}}Average Transmitted \\ Data Size (kB)\end{tabular}} & \multicolumn{4}{c|}{Transmission Latency (ms)} \\ \cline{3-6} 
                        &                                                                                               & Hu = 100 m          & Hu = 150 m         & Hu = 200 m         & Hu = 250 m        \\ \hline
Full Image              & 177.78                                                                                        &     82.384               &      82.460              &     82.537               &      82.690             \\ \hline
Ground Truth Mask              & 19.373                                                                                        &      8.973              &    8.986                &       8.994             &   9.011                \\ \hline
Segmented Mask          & 27.559                                                                                         &    12.770                &     12.782               &     12.794               &     12.818              \\ \hline
Masked Segmentation Mask& 13.728                                                                                         &     6.361               &    6.367                &     6.373               &       6.385            \\ \hline
\end{tabular}
\label{tab:Rescuenet_processing}
\end{table*}

    

\newcolumntype{C}[1]{>{\centering\arraybackslash}p{#1}} 

\begin{table*}[htbp]
\begin{center}
    \caption{Calculated Jaccard Index and MSE for the test data in FloodNet and RescueNet datasets. The matrices are calculated between the features extracted from the predicted mask and the features extracted after the product of the predicted mask with binary mask}
    \begin{tabular}{|C{4cm}|C{4cm}|C{3cm}|C{2cm}|}
    \hline
    Dataset                   & \begin{tabular}[c]{@{}c@{}}Test Data Category\end{tabular} & \begin{tabular}[c]{@{}c@{}}Jaccard Index\end{tabular} & MSE   \\ \hline
    \multirow{6}{*}{FloodNet} & Overall                                                       & 0.997                                                    & 0.016 \\ \cline{2-4} 
                              & Simple Count                                                  & 0.996                                                    & 0.018 \\ \cline{2-4} 
                              & Complex Count                                                 & 0.996                                                    & 0.019 \\ \cline{2-4} 
                              & Road Condition                                                & 0.997                                                    & 0.014 \\ \cline{2-4} 
                              & Yes/No                                                        & 0.997                                                    & 0.019 \\ \cline{2-4} 
                              & Image Condition                                               & 0.997                                                    & 0.013 \\ \hline
    RescueNet                 & Overall                                                       & 0.993                                                    & 0.029 \\ \hline
    \end{tabular}
    \label{tab:jacc}
\end{center}
\end{table*}

Table \ref{tab:jacc} presents comparisons of the calculated Jaccard index and Mean Square Error for different types of questions in the FloodNet dataset 
and damage classifications for the RescueNet dataset. For the Jaccard Index calculation in the VQA task, the normalized feature values are taken from the last layer of the visual encoder. For the building damage classification task, the feature map after the initial convolution layer is used. In both instances, the predicted mask and the relevant product of the binary mask with the predicted semantic segmentation mask are used as two sets for the calculation. When analyzing Table \ref{tab:jacc} we see Jaccard indexes lying close to 1 for all the question types in the FloodNet dataset and the classification task in the RescueNet dataset. This clearly indicates that the binary mask has preserved the critical information that is required for decision-making and discarded the uninformative content. Furthermore, the semantic fidelity is calculated as the mean square error of the normalized two vectors coming out from the last layer of the models when the predicted mask and the masked segmentation mask are in the model. The MSE values, which are close to zero depict higher similarity between two vectors implying that the binary masking is good at preserving critical data for decision making.  
\begin{table*}[!t]
\begin{center}
\caption{Parameters of model architectures.}
\label{tab:onboard}
\begin{tabular}{|l|l|c|c|c|c|}
\hline
\multicolumn{1}{|c|}{Dataset} & \multicolumn{1}{c|}{Model}  & 
\begin{tabular}[c]{@{}c@{}}Parameters\\ (Millions)\end{tabular} & \begin{tabular}[c]{@{}c@{}}FLOPs\\ (GFLOPS)\end{tabular} & \begin{tabular}[c]{@{}c@{}}MACs\\ (GMACs)\end{tabular} \\ \hline
\multirow{2}{*}{FloodNet}& Semantic Segmentation                                                                           & 49.073                                                          & 70.661                                                   & 35.286                                                 \\ \cline{2-5} 
                              & Data Masking                                                                                       & 0.044                                                           & 26.996                                                   & 13.359                                                 \\ \hline
\multirow{2}{*}{RescueNet}    & Semantic Segmentation                                                                           & 49.074                                                          & 75.575                                                   & 37.741                                                 \\ \cline{2-5} 
                              & Data Masking                                                                                     & 0.044                                                           & 27.238                                                   & 13.478                                                 \\ \hline
\end{tabular}
\end{center}
\end{table*}

Moreover, the number of parameters, floating-point operations (FLOPs), and multiply-accumulate operations (MACs) of the semantic segmentation model and FCN-based mask prediction model are evaluated for both FloodNet and RescueNet. The results are shown in Table \ref{tab:onboard} which indicate that the integration of the data masking model in the proposed framework does not add a significant overhead to the overall framework. 

\section{Conclusion}
In this paper, we proposed a semantic information extractor which is formulated as a binary mask-predicting model that can filter the data that needs to be transmitted to the ground station. Our framework enables efficient and timely decision-making in disaster monitoring and can replace existing approaches that do not pay attention to the downstream task at hand. Our experiments, conducted using two diverse downstream tasks: VQA and Image classification using two different datasets, namely, FloodNet and RescueNet, demonstrate the task-agnostic nature of our approach. In both tasks, the accuracy of the downstream task remains relatively consistent while simultaneously achieving a significant reduction in the amount of data to be transmitted. As a future task, we will investigate applications of the proposed model in satellite-to-satellite communication and satellite-to-ground communication. 
\bibliographystyle{IEEEtran} 
\bibliography{bibfile}

\begin{thebibliography}{10}
\providecommand{\url}[1]{#1}
\csname url@samestyle\endcsname
\providecommand{\newblock}{\relax}
\providecommand{\bibinfo}[2]{#2}
\providecommand{\BIBentrySTDinterwordspacing}{\spaceskip=0pt\relax}
\providecommand{\BIBentryALTinterwordstretchfactor}{4}
\providecommand{\BIBentryALTinterwordspacing}{\spaceskip=\fontdimen2\font plus
\BIBentryALTinterwordstretchfactor\fontdimen3\font minus \fontdimen4\font\relax}
\providecommand{\BIBforeignlanguage}[2]{{%
\expandafter\ifx\csname l@#1\endcsname\relax
\typeout{** WARNING: IEEEtran.bst: No hyphenation pattern has been}%
\typeout{** loaded for the language `#1'. Using the pattern for}%
\typeout{** the default language instead.}%
\else
\language=\csname l@#1\endcsname
\fi
#2}}
\providecommand{\BIBdecl}{\relax}
\BIBdecl

\bibitem{WMO_2021}
{World Meteorological Organization (WMO)}, ``{WMO Atlas of mortality and economic losses from weather, climate, and water extremes (1970–2019)},'' 2021.

\bibitem{gar2023}
{United Nations Office for Disaster Risk Reduction (UNDRR)}, ``{GAR} special report 2023: Mapping resilience for the sustainable development goals,'' 2023.

\bibitem{karaman2024solutions}
B.~Karaman, I.~Basturk, S.~Taskin, E.~Zeydan, F.~Kara, E.~A. Beyazit, M.~Camelo, E.~Bj{\"o}rnson, and H.~Yanikomeroglu, ``Solutions for sustainable and resilient communication infrastructure in disaster relief and management scenarios,'' \emph{arXiv preprint arXiv:2410.13977}, 2024.

\bibitem{lou2016onboard}
Y.~Lou, D.~Clark, P.~Marks, R.~J. Muellerschoen, and C.~C. Wang, ``Onboard radar processor development for rapid response to natural hazards,'' \emph{IEEE J. Sel. Topics Appl. Earth Observ. Remote Sens.}, vol.~9, no.~6, pp. 2770--2776, 2016.

\bibitem{alsamhi2022uav}
S.~H. Alsamhi, A.~V. Shvetsov, S.~Kumar, S.~V. Shvetsova, M.~A. Alhartomi, A.~Hawbani, N.~S. Rajput, S.~Srivastava, A.~Saif, and V.~O. Nyangaresi, ``{UAV} computing-assisted search and rescue mission framework for disaster and harsh environment mitigation,'' \emph{Drones}, vol.~6, no.~7, p. 154, 2022.

\bibitem{Ijaz_2023_UAV}
H.~Ijaz, R.~Ahmad, R.~Ahmed, W.~Ahmed, Y.~Kai, and W.~Jun, ``A uav-assisted edge framework for real-time disaster management,'' \emph{IEEE Transactions on Geoscience and Remote Sensing}, vol.~61, pp. 1--13, 2023.

\bibitem{Raja_2024_UGEN}
G.~Raja, A.~Manoharan, and H.~Siljak, ``Ugen: Uav and gan-aided ensemble network for post-disaster survivor detection through oran,'' \emph{IEEE Transactions on Vehicular Technology}, vol.~73, no.~7, pp. 9296--9305, 2024.

\bibitem{Luo_WCM_2022}
X.~Luo, H.-H. Chen, and Q.~Guo, ``Semantic communications: {O}verview, open issues, and future research directions,'' \emph{IEEE Wirel. Commun.}, vol.~29, no.~1, pp. 210--219, 2022.

\bibitem{lokumarambage2023wireless}
M.~U. Lokumarambage, V.~S.~S. Gowrisetty, H.~Rezaei, T.~Sivalingam, N.~Rajatheva, and A.~Fernando, ``Wireless end-to-end image transmission system using semantic communications,'' \emph{IEEE Access}, vol.~11, pp. 37\,149--37\,163, 2023.

\bibitem{semanticsegment_yu2018semantic}
B.~Yu, L.~Yang, and F.~Chen, ``Semantic segmentation for high spatial resolution remote sensing images based on convolution neural network and pyramid pooling module,'' \emph{IEEE J. Sel. Topics Appl. Earth Observ. Remote Sens.}, vol.~11, no.~9, pp. 3252--3261, 2018.

\bibitem{semanticmaskreiher2020sim2real}
L.~Reiher, B.~Lampe, and L.~Eckstein, ``A sim2real deep learning approach for the transformation of images from multiple vehicle-mounted cameras to a semantically segmented image in bird’s eye view,'' in \emph{Proc. IEEE 23rd Intl. Conf. Intelligent Transportation Systems (ITSC)}, 2020, pp. 1--7.

\bibitem{Lee_2024_UAVondevice}
H.~Lee, G.~Kim, S.~Ha, and H.~Kim, ``Lightweight disaster semantic segmentation for uav on-device intelligence,'' in \emph{IGARSS 2024 - 2024 IEEE International Geoscience and Remote Sensing Symposium}, 2024, pp. 8821--8825.

\bibitem{Floodnet}
M.~Rahnemoonfar, T.~Chowdhury, A.~Sarkar, D.~Varshney, M.~Yari, and R.~R. Murphy, ``Floodnet: {A} high resolution aerial imagery dataset for post flood scene understanding,'' \emph{IEEE Access}, vol.~9, pp. 89\,644--89\,654, 2021.

\bibitem{Rescuenet}
M.~Rahnemoonfar, T.~Chowdhury, and R.~Murphy, ``Rescuenet: {A} high resolution {UAV} semantic segmentation dataset for natural disaster damage assessment,'' \emph{Scientific Data}, vol.~10, no.~1, p. 913, 2023.

\bibitem{Alotaibi_2019_SAR}
E.~T. Alotaibi, S.~S. Alqefari, and A.~Koubaa, ``Lsar: Multi-uav collaboration for search and rescue missions,'' \emph{IEEE Access}, vol.~7, pp. 55\,817--55\,832, 2019.

\bibitem{Ezequiel_2014_UAVapplications}
C.~A.~F. Ezequiel, M.~Cua, N.~C. Libatique, G.~L. Tangonan, R.~Alampay, R.~T. Labuguen, C.~M. Favila, J.~L.~E. Honrado, V.~Caños, C.~Devaney, A.~B. Loreto, J.~Bacusmo, and B.~Palma, ``Uav aerial imaging applications for post-disaster assessment, environmental management and infrastructure development,'' in \emph{2014 International Conference on Unmanned Aircraft Systems (ICUAS)}, 2014, pp. 274--283.

\bibitem{Kyrkou_2024_UAV}
C.~Kyrkou and T.~Theocharides, ``Deep-learning-based aerial image classification for emergency response applications using unmanned aerial vehicles,'' in \emph{2019 IEEE/CVF Conference on Computer Vision and Pattern Recognition Workshops (CVPRW)}, 2019, pp. 517--525.

\bibitem{Dong_2021_SAR}
J.~Dong, K.~Ota, and M.~Dong, ``Uav-based real-time survivor detection system in post-disaster search and rescue operations,'' \emph{IEEE Journal on Miniaturization for Air and Space Systems}, vol.~2, no.~4, pp. 209--219, 2021.

\bibitem{Merkle_2023_DronesForGood}
N.~Merkle, R.~Bahmanyar, C.~Henry, S.~M. Azimi, X.~Yuan, S.~Schopferer, V.~Gstaiger, S.~Auer, A.~Schneibel, M.~Wieland, and T.~Kraft, ``Drones4good: Supporting disaster relief through remote sensing and ai,'' in \emph{2023 IEEE/CVF International Conference on Computer Vision Workshops (ICCVW)}, 2023, pp. 3772--3776.

\bibitem{alisjahbana2024deepdamagenet}
I.~Alisjahbana, J.~Li, Ben, Strong, and Y.~Zhang, ``Deepdamagenet: A two-step deep-learning model for multi-disaster building damage segmentation and classification using satellite imagery,'' 2024.

\bibitem{Wang_2022_RSimages}
W.~Wang, Y.~Chen, and P.~Ghamisi, ``Transferring cnn with adaptive learning for remote sensing scene classification,'' \emph{IEEE Transactions on Geoscience and Remote Sensing}, vol.~60, pp. 1--18, 2022.

\bibitem{Kyrkou_2019_ERNet}
C.~Kyrkou and T.~Theocharides, ``Deep-learning-based aerial image classification for emergency response applications using unmanned aerial vehicles,'' in \emph{2019 IEEE/CVF Conference on Computer Vision and Pattern Recognition Workshops (CVPRW)}, 2019, pp. 517--525.

\bibitem{Rashid2024TinyVQACM}
\BIBentryALTinterwordspacing
H.-A. Rashid, A.~Sarkar, A.~Gangopadhyay, M.~Rahnemoonfar, and T.~Mohsenin, ``Tiny{VQA}: {C}ompact multimodal deep neural network for visual question answering on resource-constrained devices,'' in \emph{Proc. tinyML Research Symposium – 2024}, 2024. [Online]. Available: \url{https://arxiv.org/abs/2404.03574}
\BIBentrySTDinterwordspacing

\bibitem{Kyrkou_2020_EmergencyNet}
C.~Kyrkou and T.~Theocharides, ``Emergencynet: Efficient aerial image classification for drone-based emergency monitoring using atrous convolutional feature fusion,'' \emph{IEEE Journal of Selected Topics in Applied Earth Observations and Remote Sensing}, vol.~13, pp. 1687--1699, 2020.

\bibitem{10126467}
S.~Bhorge, M.~Rane, N.~Rane, M.~Patil, P.~Saraf, and J.~Nilgar, ``Visual {AI} for satellite imagery perspective: {A} visual question answering framework in the geospatial domain,'' in \emph{Proc. 2023 IEEE 8th International Conference for Convergence in Technology (I2CT)}, 2023, pp. 1--6.

\bibitem{sarkar_chowdhury_murphy_gangopadhyay_rahnemoonfar_2023}
A.~Sarkar, T.~Chowdhury, R.~R. Murphy, A.~Gangopadhyay, and M.~Rahnemoonfar, ``{SAM}-{VQA}: {S}upervised attention-based visual question answering model for post-disaster damage assessment on remote sensing imagery,'' \emph{IEEE Trans. Geosci. Remote Sens.}, vol.~61, pp. 1--16, 2023.

\bibitem{9151332}
W.~Kim, A.~Kanezaki, and M.~Tanaka, ``Unsupervised learning of image segmentation based on differentiable feature clustering,'' \emph{IEEE Trans. Image Process.}, vol.~29, pp. 8055--8068, 2020.

\bibitem{9756908}
S.~W. Zamir, A.~Arora, S.~Khan, M.~Hayat, F.~S. Khan, M.-H. Yang, and L.~Shao, ``Learning enriched features for fast image restoration and enhancement,'' \emph{IEEE Trans. Pattern Anal. Mach. Intell.}, vol.~45, no.~2, pp. 1934--1948, 2023.

\bibitem{deepSCweng2021semantic}
Z.~Weng and Z.~Qin, ``Semantic communication systems for speech transmission,'' \emph{IEEE J. Sel. Areas Commun.}, vol.~39, no.~8, pp. 2434--2444, 2021.

\bibitem{contextbasedsemanticzhang2022context}
Y.~Zhang, H.~Zhao, J.~Wei, J.~Zhang, M.~F. Flanagan, and J.~Xiong, ``Context-based semantic communication via dynamic programming,'' \emph{IEEE Trans. Cogn. Commun. Netw.}, vol.~8, no.~3, pp. 1453--1467, 2022.

\bibitem{hu_wu_wu_xiong_2019}
Q.~Hu, C.~Wu, Y.~Wu, and N.~Xiong, ``{UAV} image high fidelity compression algorithm based on generative adversarial networks under complex disaster conditions,'' \emph{IEEE Access}, vol.~7, pp. 91\,980--91\,991, 2019.

\bibitem{https://doi.org/10.48550/arxiv.2407.05007}
V.~Polushko, A.~Jenal, J.~Bongartz, I.~Weber, D.~Hatic, R.~Rösch, T.~März, M.~Rauhut, and A.~Weinmann, ``Blessemflood21: {A}dvancing flood analysis with a high-resolution georeferenced dataset for humanitarian aid support,'' in \emph{Proc. 2024 IEEE International Geoscience and Remote Sensing Symposium (IGARSS)}, 2024, pp. 368--373.

\bibitem{PSPNetzhao2017pyramid}
H.~Zhao, J.~Shi, X.~Qi, X.~Wang, and J.~Jia, ``Pyramid scene parsing network,'' in \emph{Proc. 2017 IEEE Conference on Computer Vision and Pattern Recognition (CVPR)}, 2017, pp. 6230--6239.

\bibitem{DBLP:journals/corr/HeZRS15}
K.~He, X.~Zhang, S.~Ren, and J.~Sun, ``Deep residual learning for image recognition,'' in \emph{Proc. 2016 IEEE Conference on Computer Vision and Pattern Recognition (CVPR)}, 2016, pp. 770--778.

\bibitem{kane_khose_2022}
\BIBentryALTinterwordspacing
A.~Kane and S.~Khose, ``An efficient modern baseline for {F}lood{N}et {VQA},'' in \emph{Proc. ICML 2022 New In Machine Learning (NewInML) Workshop}, 2022. [Online]. Available: \url{https://arxiv.org/pdf/2205.15025}
\BIBentrySTDinterwordspacing

\bibitem{Liu2019RoBERTaAR}
\BIBentryALTinterwordspacing
Y.~Liu, M.~Ott, N.~Goyal, J.~Du, M.~Joshi, D.~Chen, O.~Levy, M.~Lewis, L.~Zettlemoyer, and V.~Stoyanov, ``Ro{BERT}a: {A} robustly optimized {BERT} pretraining approach,'' 2019. [Online]. Available: \url{https://arxiv.org/abs/1907.11692}
\BIBentrySTDinterwordspacing

\bibitem{qi2024minimizing}
S.~Qi, B.~Lin, Y.~Deng, X.~Chen, and Y.~Fang, ``Minimizing maximum latency of task offloading for multi-{UAV}-assisted maritime search and rescue,'' \emph{IEEE Trans. Veh. Technol.}, vol.~73, no.~9, pp. 13\,625--13\,638, 2024.

\bibitem{khan2024efficient}
N.~Khan, A.~Ahmad, A.~Wakeel, Z.~Kaleem, B.~Rashid, and W.~Khalid, ``Efficient uavs deployment and resource allocation in uav-relay assisted public safety networks for video transmission,'' \emph{IEEE Access}, vol.~12, pp. 4561--4574, 2024.

\end{thebibliography}



%








\end{document}